\documentclass[11pt]{article}
\usepackage{graphicx}
\usepackage{caption}
\usepackage{subcaption}
\usepackage{pifont}

\usepackage{natbib}
\usepackage{url}
\usepackage[final]{nips_2017}

\usepackage{amsmath}

\begin{document}

\title{GPflowOpt: A Bayesian Optimization Library \\ using TensorFlow}

\author{
	Nicolas~Knudde\thanks{ORCID ID: 0000-0001-5322-5930} \quad Joachim van der Herten \AND Tom Dhaene \quad Ivo Couckuyt\thanks{Ivo Couckuyt is a post-doctoral research fellow of FWO-Flanders.} \\
	Ghent University -- imec \\
	IDLab -- Department of Information Technology \\
	\texttt{\{nicolas.knudde, joachim.vanderherten, tom.dhaene, ivo.couckuyt\}@ugent.be}
}

\maketitle

\begin{abstract}%
A novel Python framework for Bayesian optimization known as GPflowOpt is introduced. The package is based on the popular GPflow library for Gaussian processes, leveraging the benefits of TensorFlow including automatic differentiation, parallelization and GPU computations for Bayesian optimization. Design goals focus on a framework that is easy to extend with custom acquisition functions and models. The framework is thoroughly tested and well documented, and provides scalability. The current released version of GPflowOpt includes some standard single-objective acquisition functions, the state-of-the-art max-value entropy search, as well as a Bayesian multi-objective approach. Finally, it permits easy use of custom modeling strategies implemented in GPflow.
\end{abstract}

\section{Bayesian Optimization}
Bayesian Optimization (BO) is a principled way to find a global optimum of an objective function over a bounded domain, formally expressed as
\begin{equation}
\underset{\mathbf{x} \in \mathcal{X}}{\arg \max} ~ f(\mathbf{x}).
\end{equation}
The standard configuration for BO applies the principle of dynamic programming and sequentially generates a single candidate decision $\mathbf{x}^\star$ for evaluation. Given the expensive nature of $f$, the aim is to keep the number of iterations required to identify optimal values small. All previous evaluations are used to train a (Bayesian) model which supports the search for the next decision. BO frequently applies the non-parametric Bayesian models known as Gaussian Processes (GPs) \citep{rasmussen2006gaussian} to act as a surrogate of the objective function(s). To determine the next candidate an acquisition function is maximized over the compact domain \citep{snoek2012practical}. This acquisition function usually maps the predictive distribution of the underlying model to a scalar value. 

Several extensions have been proposed to this standard setting. Batch BO evaluates batches sequentially \citep{ginsbourger2010kriging, gonzalez2016batch} to make use of parallel evaluation of the objective. The objective itself may be multi-dimensional for which multiple equally optimal solutions exist involving a trade-off (the Pareto front), this setting can be approached with multi-objective BO \citep{Couckuyt:2014}.
 
\section{Motivation}
\label{sec:motivation}
There are many libraries for BO available, whether it be commercial or open-source. Of the latter category Spearmint \citep{snoek2012practical} and GPyOpt \citep{gpyopt2016} are well-known: both packages are written in Python and rely on NumPy for numerical operations. Alternatives include BayesOpt \citep{JMLR:v15:martinezcantin14a}, implemented in C++, and RoBO \citep{robo2016}.

These available packages are written in a modular way and are relatively easy to use, but may lack extensive documentation or rigorous testing. Furthermore, adding new acquisition functions is usually quite easy but introducing a different modeling strategy requires serious effort. Confronted with the current research challenges and novel application domains for BO this motivated the development of a new framework: we intended to create an interface that is straightforward to use, such that it becomes easy to extend the code and develop new techniques with a minimum of overhead. Finally, the recent development of computing libraries supporting automated differentiation and providing scalability provide a natural base for a Bayesian optimization package.

\section{Design choices}
As the choice of modeling framework has a significant impact on the design of the resulting BO framework, several alternatives were evaluated. Ultimately we chose GPflow \citep{JMLR:v18:16-537} as framework for modeling: GPs are the most common surrogate model used in BO, and GPflow makes development and implementation of custom GP models for BO considerably easier. The package is written in Python and provides a powerful framework for implementing (GP) models, including Sparse GPs and GP-Latent Variable Models, using variational inference as the standard approximate inference technique. As GPflow is built on TensorFlow, it enables use of GPU computations, parallelization and automatic differentiation. 

The development resulted in the release of the open-source GPflowOpt project featuring following properties:
\begin{enumerate}
	\item Simple application of \textbf{different models} (using the GPflow framework) as a surrogate in Bayesian Optimization,
	\item \textbf{Automated differentiation} increasing ease of implementation,
	\item Support for (multiple) \textbf{GPU} enabling fast computation,
	\item Clean \textbf{object-oriented Python} front-end which is simple to extend,
	\item Rigorous \textbf{testing} and \textbf{extensive} documentation.
\end{enumerate}

The code is completely modular permitting simple implementation of different acquisition functions. Acquisition functions that are  included with the framework are summarized in Table \ref{tab:ac}. Both single- and multi-objective acquisition functions are implemented as well as Probability of Feasibility (PoF) to incorporate black-box constraints. GPflowOpt supports most models included in GPflow such as the Variational Gaussian Process (VGP) or the Sparse Variational Gaussian Process (SVGP), and allows the use of custom models. The next version of GPflow intends to further increase these capabilities as integration of other TensorFlow modeling frameworks (such as Keras) with GPflow will be possible.

\begin{table}
\setlength{\abovecaptionskip}{10pt plus 3pt minus 2pt}
\centering
\begin{tabular}{| l | l |}
	\hline
	Single-objective & Expected Improvement (EI) \citep{movckus1975bayesian} \\
	& Lower Confidence Bound (LCB)\citep{icml2010_SrinivasKKS10}\\
	& Max-Value Entropy Search (MES) \citep{pmlr-v70-wang17e} \\
	& Probability of Improvement (PoI) \citep{kushner1964new} \\ \hline
	Multi-objective & Hypervolume-based PoI (HvPoI) \citep{Couckuyt:2014} \\ \hline
	Constraint & Probability of Feasibility (PoF) \citep{Schonlau:1997} \\
	\hline
\end{tabular}
\caption{Implemented acquisition functions}
\label{tab:ac}
\end{table}

\section{Implementation}
\begin{figure}
	\centering
		\includegraphics[width=\textwidth]{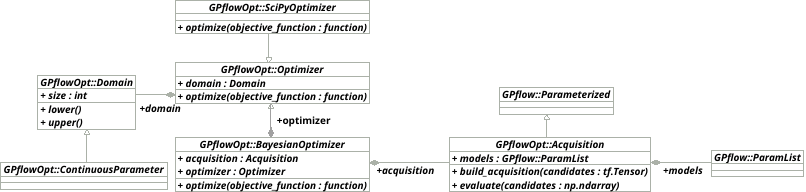}
	\caption{UML diagram of GPflowOpt}
	\label{fig:UML}
\end{figure}

Following the modular structure of GPflow, the main building blocks of GPflowOpt are \verb|Domain|, \verb|Acquisition| and \verb|Optimizer|. The class relationships are summarized in Figure \ref{fig:UML}.

The expensive objective function $f$ is defined by the user. The domain is a GPflowOpt class containing the bounds of the optimization domain, which is used for scaling purposes and configuration of optimizer objects. The acquisition function (\verb|Acquisition|) holds one or more GPflow models and maps their predictions to a score. By default a transparent model wrapper is used for automated data scaling to and from the underlying model to increase the success rate of the hyperparameter optimization. The \verb|BayesianOptimizer| class handles the classic BO process and includes model optimization, optimization of the acquisition function, evaluation of the objective and optionally marginalization of hyperparameters using Hamiltonian Monte-Carlo sampling.

\section{Comparison}
In Table~\ref{tab:comp}, a comparison is made to the popular BO frameworks summarized in Section~\ref{sec:motivation}. Most frameworks are written in Python. Key features of GPflowOpt over other frameworks include the support for multi-objective objective functions along with an implementation of an acquisition function specifically for this type of applications, the rigorous testing suite resulting in a code coverage of 99\% and extensive documentation. Additionally, a fast algorithm for generating maximin Latin hypercubes \cite{Viana:2010} is included. Latin hypercubes are a popular method for generating space-filling Design of Experiments (DoE) to start the Bayesian optimization process. 

On the other hand GPyOpt supports batch BO, which is currently still in development for GPflowOpt and was not part of the first release. In terms of efficient hardware usage a native implementation in C++ as offered by BayesOpt is preferable, however the computations in GPflowOpt are carried out by TensorFlow graphs which benefit of a native computational back-end. The additional scalability and automated differentiation compensate for the overhead of the framework. Another powerful feature of GPflowOpt is the option to implement models in the GPflow framework, allowing their use without the need of implementing wrapper classes due to the modular structure. 
\begin{table}[t]
\setlength{\abovecaptionskip}{10pt plus 3pt minus 2pt}
	\centering
	\begin{tabular}{| l | c | c | c | c| c|}
	\hline
		 & GPflowOpt & GPyOpt & Spearmint  & BayesOpt & RoBO \\ \hline
		Language & Python & Python & Python & C++ & Python \\
		Auto differentiation & \ding{51} & \ding{55} & \ding{55} & \ding{55} & \ding{55} \\
		Multi-objective & \ding{51} & \ding{55} & \ding{55} & \ding{55} & \ding{55} \\
		Code coverage & 99\% & 56\% & -- & -- & 46\% \\
		Batch BO & \ding{55} & \ding{51} & \ding{55} & \ding{55} & \ding{55} \\
		GPU support & \ding{51} & \ding{55} & \ding{55} & \ding{55} & \ding{55} \\
		\hline
	\end{tabular}
	\caption{Comparison of Bayesian optimization frameworks. The information are taken from the respective web pages or papers \citep{gpyopt2016, snoek2012practical,JMLR:v15:martinezcantin14a,robo2016}}
	\label{tab:comp}
\end{table}

\section{Illustration}
\begin{figure}[t]
	\centering
	\begin{subfigure}[b]{0.49\textwidth}
		\includegraphics[width=\textwidth]{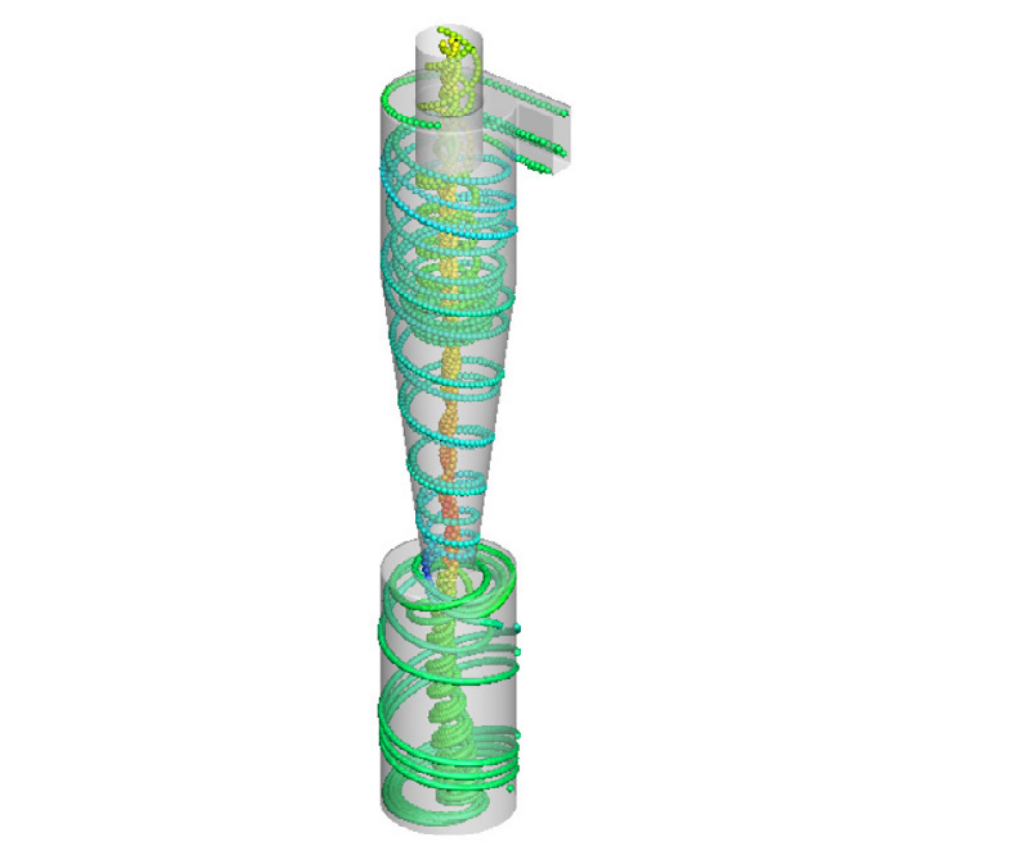}
		\caption{Cyclone}
		\label{fig:cycl}
	\end{subfigure}
	\begin{subfigure}[b]{0.49\textwidth}
		\includegraphics[width=\textwidth]{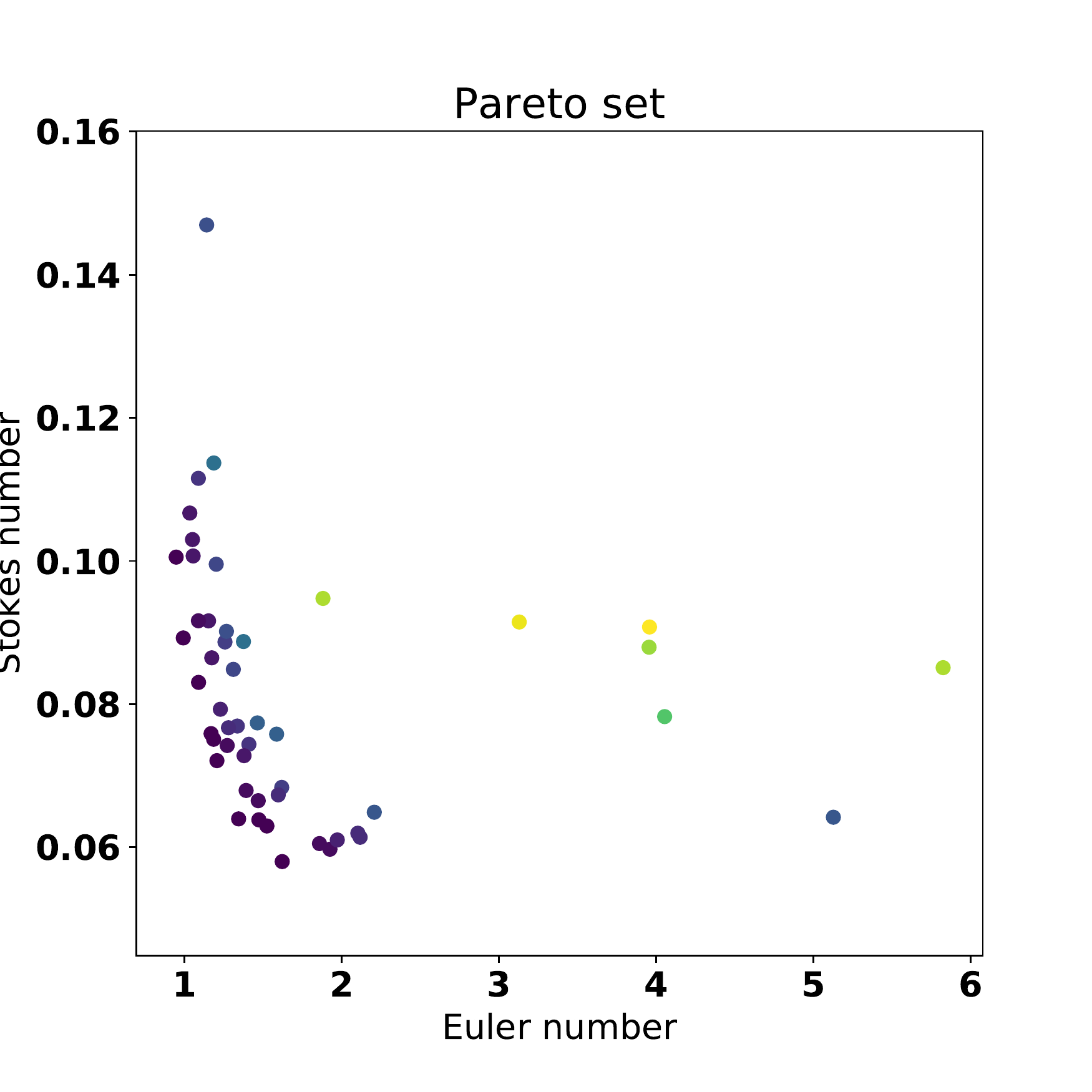}
		\caption{Resulting Pareto front}
		\label{fig:pareto}
	\end{subfigure}
	\caption{Multi-objective optimization of a cyclone separator (shown on the left). The Pareto front only depicts the feasible samples in the objective space. The bottom-left corner represents the optimal position, and the colors indicate its dominance.}
\end{figure}

The following example presents the optimization of a gas cyclone separator (depicted in Figure~\ref{fig:cycl}), a real-world device that is able to separate dust particles from gases through a complex swirling motion. Finding an optimal solution involves a trade-off between two conflicting objectives which will be solved with multi-objective BO. The device is characterized by 7 geometric parameters, which are the input of the expensive objective function which results in two objectives: the pressure loss (represented by the Euler number) and the cut-off diameter (represented by the Stokes number). The multi-objective BO approach used here is the Hypervolume Probability of Improvement (HvPoI) \citep{Couckuyt:2014}, which indicates the probability of a candidate evaluation improving the volume between the Pareto front and a reference point (e.g., the anti-ideal point) in the objective space.
At the same time four production inequality constraints based on the same inputs have to be taken into account. These constraints are black-box themselves and are modeled as well. By incorporating the Probability of Feasibility (PoF) \citep{Schonlau:1997} into a joint acquisition function this aspect can be included, as the feasibility is learnt jointly with the objectives. 

A maximin Latin hypercube consisting of 50 points was used as an initial design and is implemented by using the Translational Propagation algorithm \citep{Viana:2010}. A total of 120 evaluations of the objective function were performed, each evaluation yields both the Euler and Stokes numbers, as well as the constraint values. The resulting Pareto front of the feasible samples is shown in Figure~\ref{fig:pareto}.

\section{Conclusion and future work}
A new, versatile Python package for Bayesian optimization was introduced. It allows for straightforward TensorFlow model integration by building on GPflow. This offers significant benefits such as automatic differentiation, multi-core calculations and GPU support. A comparison was made with other open-source packages indicating the package offers significant advantages including rigorous testing and extensive documentation. Currently, the framework still lacks advanced sampling techniques such as batch BO. The current short-term roadmap of GPflowOpt is batch BO as well as support for discrete and categorical variables.

The development of GPflowOpt is open-source and fully transparent. Hence, the scientific community is encouraged to make contributions to the framework and test out their own Bayesian optimization algorithms.

\bibliography{bib}
\bibliographystyle{apalike}
\end{document}